\documentclass[10pt,twocolumn,letterpaper]{article}
\usepackage[pagenumbers]{iccv}
\usepackage{mmstyle}

\PassOptionsToPackage{table,dvipsnames}{xcolor}
\usepackage{xcolor}

\usepackage{colortbl}
\usepackage{graphicx}
\usepackage{booktabs}
\usepackage{multirow}

\newcommand{\myPara}[1]{\noindent\textbf{#1}}

\newcommand{\fk}[1]{{\color{black}#1}}
\usepackage{hyperref}

\def\paperID{1838}
\def\confName{ICCV}
\def\confYear{2025}

\title{Go to Zero: Towards Zero-shot Motion Generation with Million-scale Data}

\author{Ke Fan\\
Shanghai Jiao Tong University\\
\and
Shunlin Lu\\
CUHK, Shenzhen\\
\and
Minyue Dai\\
Fudan University\\
\and
Runyi Yu\\
HKUST\\
\and
Lixing Xiao\\
Zhejiang University\\
\and
Zhiyang Dou\\
HKU\\
\and
Junting Dong\\
Shanghai AI Laboratory\\
\and
Lizhuang Ma\textsuperscript{$\dagger$} \\
Shanghai Jiao Tong University, East China Normal University\\
\and
Jingbo Wang\\
Shanghai AI Laboratory\\
}

\begin{document}
\twocolumn[{%
\renewcommand\twocolumn[1][]{#1}%
\maketitle
\vspace{-3em}
\begin{center}
    \centering
    \captionsetup{type=figure}
    \includegraphics[width=\textwidth]{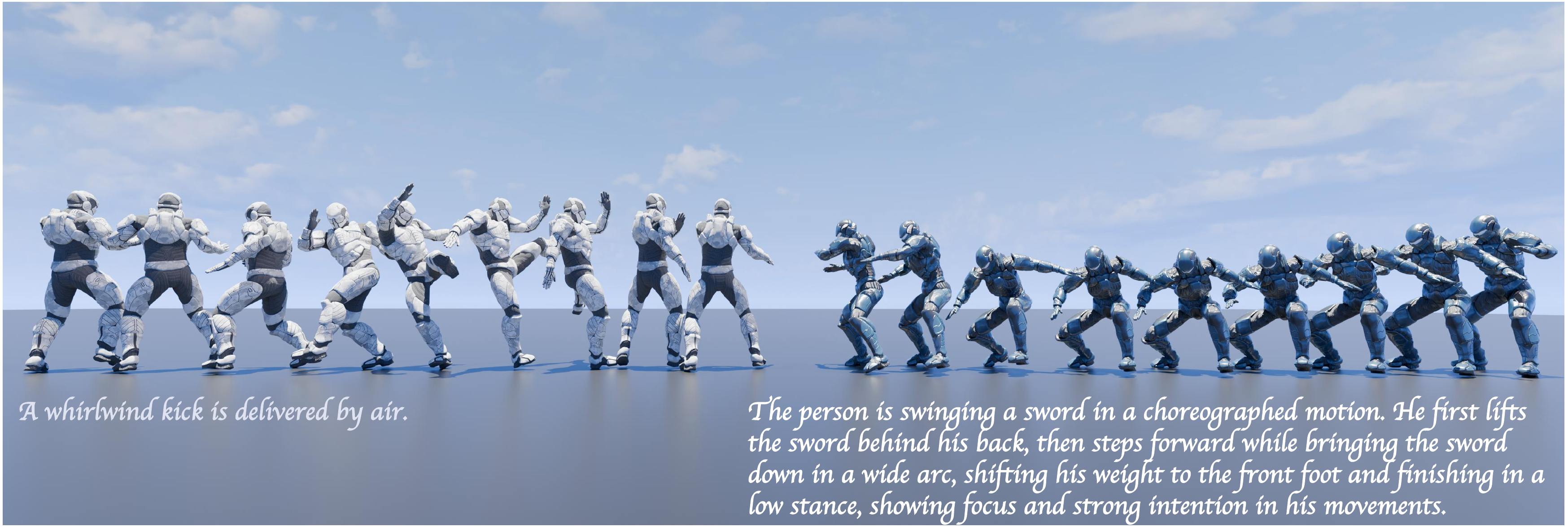}
    \captionof{figure}{We present Go to Zero, where we can deal with out-domain and complex compositional motions.}
    \label{fig:teaser}
\end{center}
}]
\let\thefootnote\relax\footnotetext{$^\dagger$ Corresponding author}

\begin{abstract}
Generating diverse and natural human motion sequences based on textual descriptions constitutes a fundamental and challenging research area within the domains of computer vision, graphics, and robotics. Despite significant advancements in this field, current methodologies often face challenges regarding zero-shot generalization capabilities, largely attributable to the limited size of training datasets. Moreover, the lack of a comprehensive evaluation framework impedes the advancement of this task by failing to identify directions for improvement. In this work, we aim to push text-to-motion into a new era, that is, to achieve the generalization ability of zero-shot. To this end, firstly, we develop an efficient annotation pipeline and introduce MotionMillion—the largest human motion dataset to date, featuring over 2,000 hours and 2 million high-quality motion sequences. Additionally, we propose MotionMillion-Eval, the most comprehensive benchmark for evaluating zero-shot motion generation. Leveraging a scalable architecture, we scale our model to 7B parameters and validate its performance on MotionMillion-Eval. Our results demonstrate strong generalization to out-of-domain and complex compositional motions, marking a significant step toward zero-shot human motion generation. The code is available at \it{\url{https://github.com/VankouF/MotionMillion-Codes}}.
\end{abstract}    
\section{Introduction}
\label{sec:introduction}
Text-to-motion generation, which synthesizes diverse and temporally coherent human motions from natural language descriptions, poses a significant challenge in computer vision, graphics, and robotics. Despite notable advancements~\citep{rombach2022high,gpt3,gpt4, yang2025sigman, li2024dispose, chen2025dancetogether, sora, cheng2024learning, xu2023survey} in large-scale generative models for text, images, 3D, and videos—showcasing exceptional zero-shot capabilities—the motion domain~\citep{mdm,t2mgpt,zhou2023let,actformer,momask} remains considerably behind. This discrepancy is attributed not to a lack of algorithmic innovation, but to inherent limitations in data scale and model architecture that hinder robust generalization, thereby constraining real-world applicability.

To address the generalization challenges faced by state-of-the-art methods such as MDM~\citep{mdm}, MotionGPT~\citep{motiongpt}, and MoMask~\citep{momask}, which are limited by dataset constraints in HumanML3D~\citep{humanml3d} and MotionX~\citep{motionx}, earlier approaches~\citep{motionclip, avatarclip} aligned motion sequences with the image embedding space of CLIP~\citep{clip} using rendered frames. While these methods somewhat alleviate data scarcity, they suffer from inherent modality mismatches, as static image embeddings fail to capture temporal dynamics, resulting in incoherent motions and limited compositional reasoning. Recent efforts~\citep{lu2024scamo, xu2024motionbank, omg, wang2024quo} have aimed to enhance generalization by scaling up motion generation models with larger datasets (e.g., 250 hours of motion in ScaMo~\cite{lu2024scamo} and one billion parameters in OMG~\cite{omg}). These methods exhibit improved motion diversity and novel language alignment compared to previous methods~\cite{mdm, motiongpt, momask}, which were trained on more restricted datasets~\cite{humanml3d, motionx}. Nonetheless, impeded by constraints related to model capacity and the inherent limitations of the datasets (\eg quality, diversity, and size), the full potential of this scaling-up formulation remains largely unexploited, particularly when these text inputs entail long-term motion compositions and complex descriptions.

Therefore, we contend that achieving human-level motion generation necessitates a paradigm shift akin to the \emph{``scaling hypothesis"}: sufficiently large and diverse \fk{high quality} training data, combined with scaled model architectures, can unlock emergent zero-shot capabilities\fk{, especially performing the complex compositional motions}. 
\fk{To this end, in this paper we explore three key components for zero-shot motion generation: 1) large-scale and high-quality motion dataset, 2) scalable and model architecture, 3) effective evaluation benchmark.}

\fk{Firstly,} we introduce MotionMillion, a large-scale motion dataset with comprehensive text annotations. We propose a novel and efficient 
\fk{motion annotation mechanism, including motion descriptions and high quality motion capturing,} sourced from web-scale human motion videos. Our framework autonomously harvests human motion from unlabeled videos through kinematic regression, generates semantically rich captions using advanced vision-language models (\eg, gpt-4o~\cite{gpt4}), and implements a multi-stage filtering process to eliminate scene cuts and static pose jitters. This meticulous curation results in a dataset comprising \fk{over} \textbf{2000 hours} of high-quality text-motion pairs, encompassing over \textbf{2 million} motion sequences—\textbf{20 times} larger than existing resources. By unifying annotations across extant datasets (\eg, HumanML3D~\citep{humanml3d}, MotionX~\cite{motionx}), we establish a temporally coherent and compositionally diverse foundation for scaling. 

Leveraging this dataset, we \fk{further explore the effective scalable model to accommodate our large-scale dataset. LLAMA, as a transformer decoder-only model, has significantly demonstrated its scalability in text generation tasks. In the motion field, it has been proven in Scamo~\citep{lu2024scamo} that a scaling law curve can be drawn with the increase in data volume. Therefore, 1) we first use Finite Scalar Quantization (FSQ) as \textbf{Efficient Motion Tokenization:} to discretely encode the motion data. This is a more stable and efficient way compared to VQ-VAE. However, we found that although the discretization method of FSQ can effectively encode motion data when the data scale is limited, due to the extremely large scale of our data, directly using the FSQ model will cause jitter in the reconstructed motion. We believe that this is because the information loss caused by the discretization method of FSQ becomes more serious as the data scale increases, leading the model to wrongly model high-frequency information. To solve this problem, we propose to use wavelet transformation to preprocess the motion data before inputting it into the reconstruction network. After obtaining the decoder outputs, we further utilize the inverse wavelet transformation to obtain the final reconstructed motions. By this means, we can effectively encode the motion data and reduce the jitter phenomenon caused by discretive information loss. After completing the discrete compression encoding, we further utilize the LLAMA architecture to implement}
2) \textbf{Scalable Motion Generation:} We
design a bidirectional transformer that jointly models text-motion cross-attention and autoregressive motion token prediction, enabling compositional motion synthesis. Starting from a 1B parameter base, we progressively scale model depth to the final 7B scale, observing emergent zero-shot capabilities. As shown in Fig.~\ref{fig:teaser}, our model could deal with 
\fk{various texts, especially can follow complex long texts.}

To systematically assess the zero-shot generalization capabilities of models, we \fk{further} introduce MotionMillion-Eval, a new benchmark comprising 126 diverse prompts across 7 categories, ranging from daily life scenarios to inhuman motions. Our evaluation focuses on three key aspects: text-motion alignment, motion smoothness, and physical feasibility of motions. Our findings indicate that the 7 billion parameter model, MotionMillion, successfully trained on the MotionMillion dataset, exhibits robust zero-shot generalization abilities. This advancement paves the way for advancing the motion generation task towards zero-shot applications. 

Our contributions can be summarized as follows:
\begin{itemize}
    \item We propose a high-quality annotation pipeline of human motions from video data and build MotionMillion, a large-scale human motion dataset, which is currently the largest human motion with the highest quality, and its scale and diversity drive the research in the field of human motion towards zero-shot application.
    \item We \fk{propose to leverage the wavelet transformation to descrease the jitter phenomonen from FSQ, and we further scale our model to} 7B parameters via an effective scalable architecture, demonstrating strong generalization for out-of-domain complex compositional motions.
    \item We built the MotionMillion-Eval benchmark according to industry standards, which is the first proposed evaluation that can be used for zero-shot capability verification.
\end{itemize}

\section{Related Work}
\label{sec:related_work}
\myPara{Text-aligned Human Motion Generation}~\citep{plappert2018learning, text2action, dvgans, jl2p, t2g, motionclip, temos, avatarclip, tm2t, motiondiffuse, teach, mdm, humanise, mld, mofusion, physdiff, yang2024egochoir, t2mgpt, diffprior, remodiffuse, gmd, motionclr, motiongpt, motiongptv2, unihsi, omnicontrol, humantomato, tlcontrol, momask, promotion, amd, b2ahdm, emdm, stmc, flowmdm, move, stablemofusion, zhang2024generative, ji2023stylevr, ji2025sport, Ji_2025_CVPR, xu2023actformer, xu2024regennet, dai2024motionlcm, xiao2025motionstreamer, fan2024freemotion, dai2025towards, feng2024motionwavelet, wan2024tlcontrol, zhou2024emdm, chen2024pay, cong2024laserhuman, yu2025hero} has progressed rapidly in recent years, benefiting substantially from advances in generative models~\citep{ddim, ddpm1, ddpm2, attention} and the expansion of large-scale datasets~\citep{motionx,interx,humanml3d}. 
Although physics-based motion generation methods~\citep{huang2025modskill, pan2025tokenhsi, dou2023c, pan2024synthesizing, wang2025sims, yu2025skillmimic, wang2025skillmimic} can generate actions that are more in line with physical laws, text-align kinematic motion generation can possess higher flexibility.
Methodologically, the introduction of GPT-like approaches~\citep{t2mgpt, humantomato, momask, motiongpt} and diffusion-based methods~\citep{mdm, motiondiffuse, remodiffuse,emdm,mld,motionlcm} has substantially driven innovation in human motion generation. Meanwhile, KIT~\citep{kit} and HumanML3D~\citep{humanml3d} have emerged as key benchmarks supporting text-driven motion generation. Nevertheless, the models tends to overfit the above datasets and lose the generalization capabilities. 

\myPara{Large Motion Model.} Recent studies~\citep{motionx,interx,omg} seek to enhance generation quality by scaling dataset sizes. Concurrently, various works~\citep{motiongpt, motiongptv2, wu2024motionllm, avatargpt, wang2024quo, zhang2024large} focus on enlarging model capacities, such as fine-tuning pretrained large language models~\citep{motiongpt, motiongptv2, wu2024motionllm, avatargpt, wang2024quo}, although the resulting performance has often been suboptimal. LMM~\citep{zhang2024large} implements a large diffusion model but suffers from slow training speed. ScaMo~\citep{lu2024scamo} try to scale the dataset size, motion vocabulary size, and the autoregressive model size and first to explore the scaling law in text-driven motion generation. Despite these advancements, current models exhibit limited generalization, thereby constraining their applicability in downstream tasks. We attribute this challenge primarily to the insufficient scale of current datasets. Therefore, in this work, we further expand both dataset sizes and model capacities, aiming to advance text-driven motion generation toward zero-shot scenarios.
\section{MotionMillion Construction}
\label{sec:data_construction}
\begin{figure*}[t]
    \centering
    \includegraphics[width=1\textwidth]{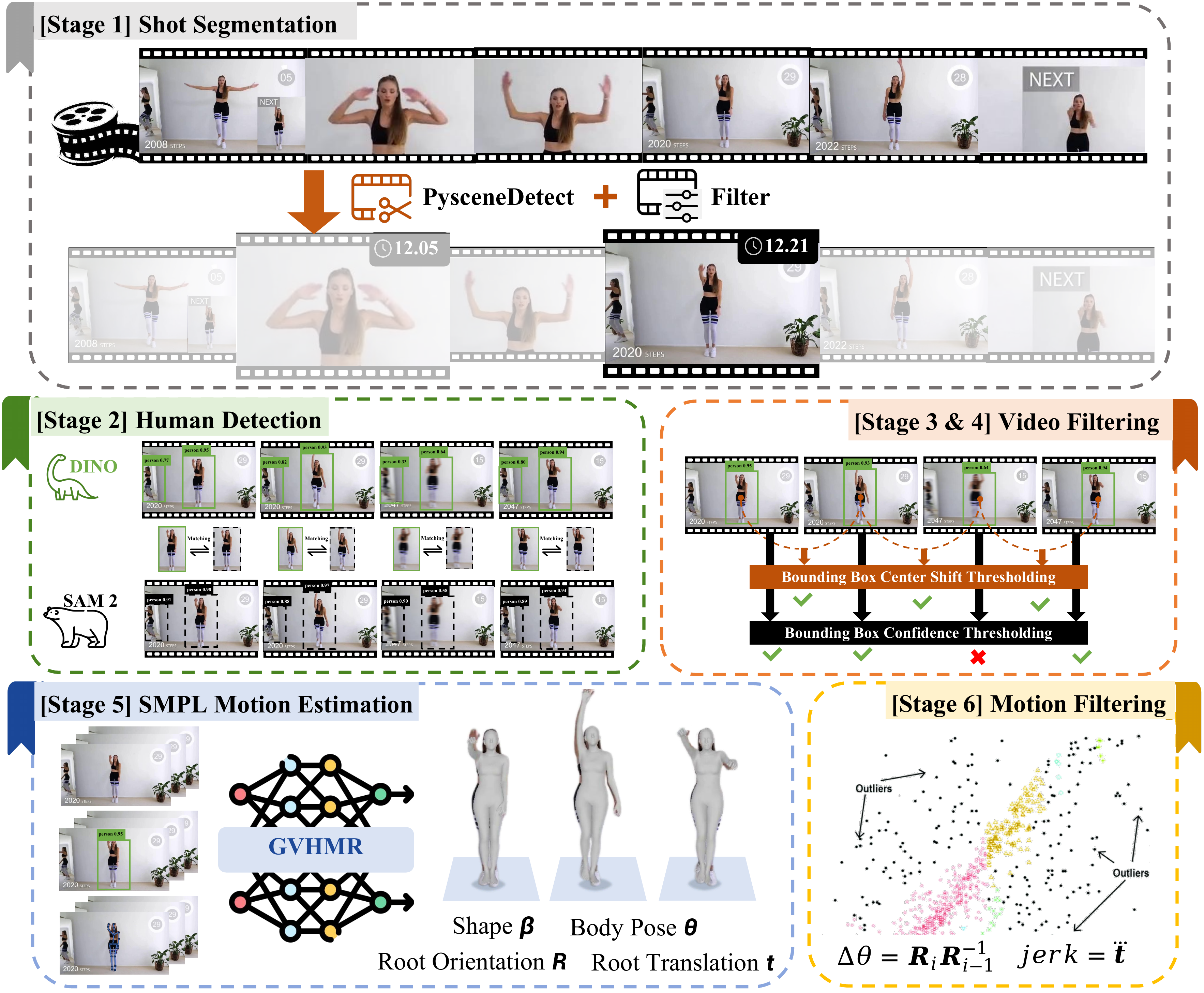}
    \vspace{-1em}
    \caption{Data Construction Pipeline of MotionMillion. We can obtain high-quality human motion from a monocular video via our six processing stages, i.e. Shot Segmentation, Human Detection, Video Filtering, SMPL Motion Estimation and Motion Filtering.}
    \vspace{-1em}
    \label{fig:data_construct}
\end{figure*}
\subsection{Overview}
Our dataset construction integrates human motion reconstruction from large-scale in-the-wild video sources and the re-aggregation of existing motion datasets. To enhance diversity and coverage, we incorporate multiple established datasets, including MotionX~\citep{motionx}, InterHuman~\citep{liang2024intergen}, Inter-X~\citep{interx}, BABEL~\citep{babel}, Fitness~\citep{motionx}, PhantomDance~\citep{danceformer}, GDance~\citep{le2023music}, FineDance~\citep{finedance}, HI4D~\citep{yin2023hi4d}, TRUMANS~\citep{jiang2024scaling}, and HumanSC3D~\citep{fieraru2021learning}. To further scale data collection, we propose an efficient pipeline for reconstructing human motion from web-scale video sources.

Our methodology primarily focuses on full-body motion while omitting hand and facial expressions. To represent human motion, we extract SMPL\citep{loper2023smpl} parameters, a widely adopted parametric model for human body articulation. As illustrated in Fig.\ref{fig:data_construct}, our motion reconstruction framework consists of six key stages: which are \textbf{I)} Shot Segmentation, \textbf{II)} Human Detection, \textbf{III)} Bounding Box Confidence Filtering, \textbf{IV)} Transition Filtering, \textbf{V)} SMPL Motion Estimation, and \textbf{VI)} Motion Filtering.
\subsection{Human Motion Reconstruction}
In this section, we will illustrate the motion reconstruction process from web-scale human motion videos in detail.

\textbf{Stage I Shot Segmentation.} 
Raw video data often exhibits varying quality and frequent scene transitions, leading to artifacts in SMPL parameter estimation and adversely impacting downstream tasks. To address these challenges, we employ PySceneDetect to segment videos into single-scene clips, enforcing temporal coherence by restricting each clip to a maximum of 200 frames. Additionally, we use the Laplacian operator from OpenCV to evaluate image sharpness, selecting the sharpest frame as the initial frame of each clip. This preprocessing pipeline enhances the robustness of motion reconstruction while mitigating noise in the extracted SMPL parameters.

\begin{figure*}[t]
    \centering
    \includegraphics[width=0.75\textwidth]{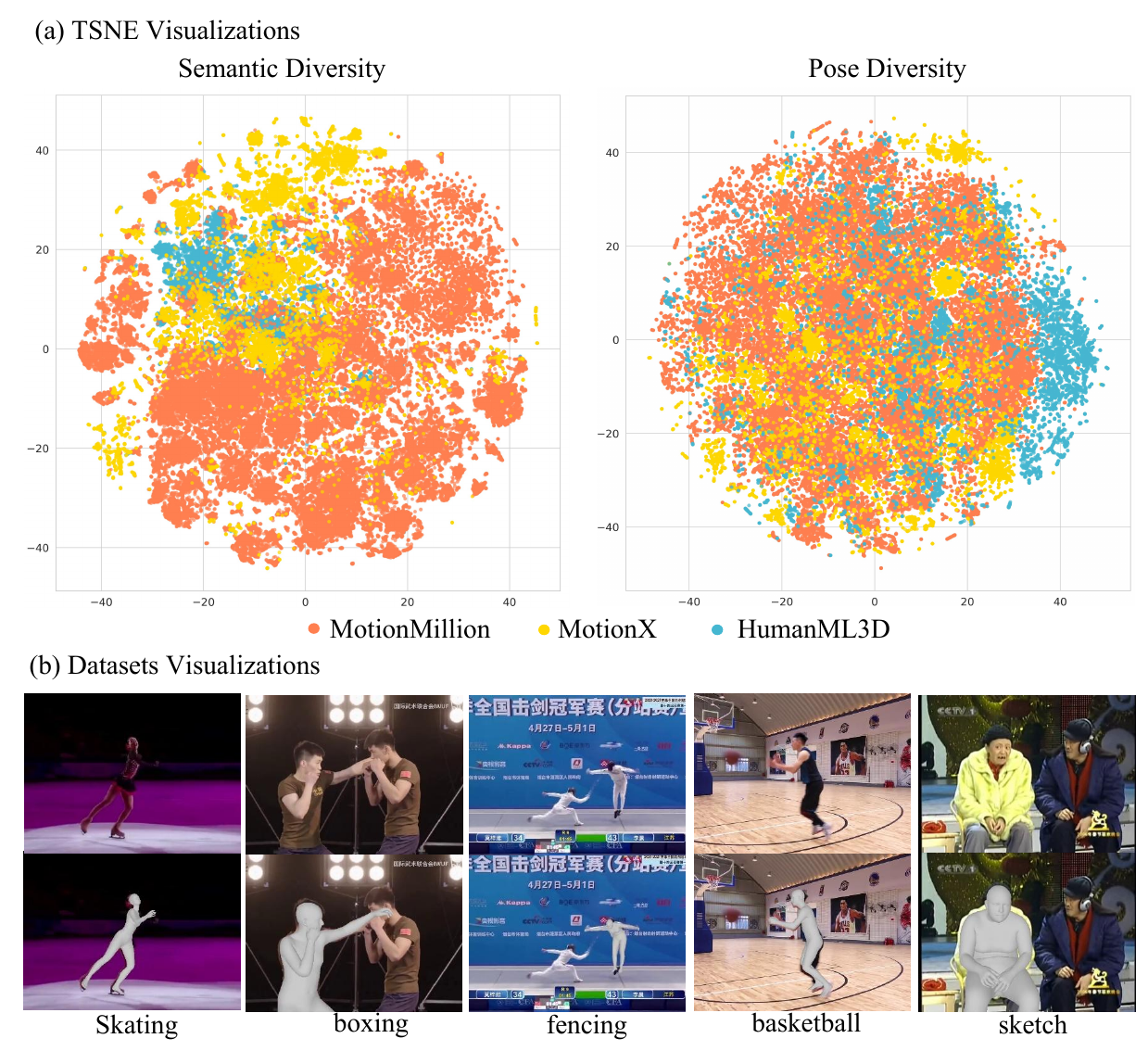}
    \vspace{-1em}
    \caption{Overview of MotionMillion. This dataset exhibits extensive semantic and pose diversity, encompassing a broad spectrum of indoor and outdoor human motions.}
    \vspace{-1em}
    \label{fig:data_analysis}
\end{figure*}

\textbf{Stage II Robust Human Detection and Tracking.} Accurate human detection and tracking are crucial for high-quality SMPL motion estimation. However, videos sourced from online platforms present two major challenges: (1) varying numbers of individuals present in each frame and (2) severe occlusions. To address these issues, we propose a \textbf{coarse-to-fine} approach that enhances the robustness of human detection and tracking, ensuring reliable motion estimation across diverse and complex scenarios. 

\textbf{During the coarse stage}, we leverage Grounding DINO~\citep{liu2023grounding} and SAM2~\citep{ravi2024sam2} to fix the problem of variable human counts.  Grounding DINO trains on an extremely large-scale dataset, and can detect any object based on the input text. SAM2 is a foundation model for solving promotable visual segmentation in images and videos. It can use the bounding box of a person in the input image as a prompt to track that person's identity information in subsequent frames. Specifically, \textbf{1)} Use Grounding DINO for high-confidence human detection. If the first frame lacks a person, scan subsequent frames until a valid detection. We empirically set the threshold at 0.85 to make sure the detected human body contains high quality. \textbf{2)} Feed the detected bounding box into SAM2, which propagates the mask across frames via prompt-based segmentation.

\textbf{During the fine stage}, we aims to solve the problem of severe occlusions.
SAM2 demonstrates strong robustness in tracking individuals, even under occlusions, successfully preserving identity information. However, occlusions significantly degrade the accuracy of human keypoint detection, thereby impacting motion reconstruction quality. To mitigate this, we leverage Grounding DINO’s confidence score as an indicator of detection reliability and introduce a refinement mechanism to filter low-quality bounding boxes: 
\textbf{1)} Calculating the IoU between SAM2 tracked boxes and Grounding DINO’s per-frame detections. \textbf{2)} Selecting candidates in Grounding DINO's detection with IoU greater than 0.85, then choosing the box with minimal area deviation from the box tracked by SAM2. \textbf{3)} If the confidence score of the selected bounding box is greater than a certain threshold, it is considered a successful match; otherwise, it is considered a failed match. By matching the corresponding bounding boxes and filtering according to the confidence level, we can effectively alleviate the situation where a large number of occluded humans are detected due to the robustness of SAM2.

\textbf{Stage III \& IV Bounding Box Confidence Filtering and Transition Filtering.} We finish the bounding box confidence filtering during the fine stage of stage 2. 
While PySceneDetect is employed for preliminary shot segmentation, it struggles with scenarios where the background remains unchanged, but the subject undergoes sudden positional shifts. To address this limitation, we introduce an additional detection step for sudden position changes. We calculate the distance between the centers of the bounding boxes of detected humans in two consecutive frames. If it is greater than a certain threshold, it is considered that there is a sudden position change, and thus the video clip is divided into two parts. 

\textbf{Stage V Human Motion Reconstruction.} To obtain high-quality human parameters, we used the GVHMR~\cite{shen2024world}, which, as the state-of-the-art method in human motion reconstruction, could recover a much more realistic motion in both camera and world space. It first propose estimating human poses per-frame in a novel Gravity-View (GV) coordinate system. The GV system is defined by the world gravity and the camera view direction, which could naturally align with gravity and largely reduce the ambiguity in defining the world coordinate system. It takes the bounding box, 2D key points, image features, and relative camera rotations as input, and leverage a transformer network to predict the SMPL parameters, including the root translation $t$, body pose $\theta$, root orientation $R$, and shape $\beta$.

\textbf{Stage VI Motion Filtering.} 
PySceneDetect algorithm is limited in detecting sudden orientation changes in the foreground, particularly in scenarios where the background remains unchanged. 
While GVHMR demonstrates strong performance in human motion estimation, it remains susceptible to errors caused by camera motion-induced jitter, leading to inconsistencies in human body estimation. Therefore, to mitigate these issues, we integrate global orientation ($R$) and joint's position ($J$) estimated by GVHMR to effectively detect sudden orientation changes while filtering out jitter-related artifacts.
To filter sudden orientation changes, we compute the transformation angle $\Delta\theta$ between two consecutive frames using the global orientation R, formulated as:
\vspace{-0.5em}
\begin{equation}
   \Delta\theta = Transform(R_iR_{i-1}^{-1})
\label{eq:orient}
\vspace{-0.5em}
\end{equation}, where $i$ indicates the index of the verified frame and the $Transform$ represents the function of transforming the rotation from matrix form into the axis-angle form. For jitter filtering, we introduce the $jerk$ metric, which is sensitive to kinetic irregularities and can effectively indicate the motion smoothness~\citep{barquero2024seamless}. The jerk is defined as the time derivative of acceleration, formulated as: 
\vspace{-0.5em}
\begin{equation}
   jerk = \dddot{J_i}
\label{eq:jerk}
\vspace{-0.5em}
\end{equation}, where $J$ represents the global position of different body joints.
After obtaining the $\Delta\theta$ and $jerk$ metrics for all of the video clips, we take the Isolation Forest~\citep{liu2008isolation} algorithm to identify the outliers for both metrics respectively, which can detect the moments when sudden orientation changes and jitters occur in the video clip in an unsupervised manner.  

Our construction mechanism can generate precise and smooth motions. Finally, we standardize each motion to 30fps according to the frame rate of the corresponding raw video, obtaining our final motion.

\subsection{Motion Caption.} Different from previous motion datasets (e.g. MotionX), our motion caption contains two steps: Motion Description and Description Augmentation. 

\textbf{Motion Description.} 
We split the video according to the results obtained from the previous human motion reconstruction to obtain the corresponding video clip. Then, we input this clip to GPT-4o to obtain the description of the human action in the corresponding bounding box. If the video title contains a description of the corresponding semantics, it is input to GPT-4o together. At the same time, different from MotionX, in addition to the description of body parts, we also focus on prompting the GPT-4o model to describe the age, body characteristics, movement styles, emotions, and environments of the subject, which is critical during motion generation. All the prompts used are shown in the appendix.

\textbf{Description Augmentation} MotionX only describes the motion once, resulting in insufficient diversity of text descriptions of the same motion and subsequently significantly inhibits the model generalization ability. Therefore, after obtaining the motion description, we further prompt the model of LLAMA 3.1-8B~\citep{llama} to rewrite the motion description 20 times without changing the original semantic meaning, so as to realize the description augmentation.
\subsection{Data Analysis}
\begin{table}[t]
\resizebox{\columnwidth}{!}{%
\begin{tabular}{lllllllll}
\hline
\multirow{2}{*}{Dataset} &
  \multirow{2}{*}{Clip Number} &
  \multirow{2}{*}{Hour} &
  \multirow{2}{*}{Text Diversity} &
  \multicolumn{1}{c}{\multirow{2}{*}{Static Motion}} &
  \multicolumn{1}{c}{\multirow{2}{*}{Person}} &
  \multicolumn{3}{c}{Scene} \\ \cline{7-9} 
                     &       &       &                  & \multicolumn{1}{c}{} & \multicolumn{1}{c}{} & Indoor & Ourdoor & RGB \\ \hline
KIT-ML               & 3911  & 11.2  & 1-3              & no                   & single               & yes    & no      & no  \\
BABEL                & 13220 & 43.5  & 1                & no                   & single               & yes    & no      & no  \\
HumanML3D            & 14616 & 28.6  & 1-3              & no                   & single               & yes    & no      & no  \\
Motion-X             & 81084 & 144.2 & 1                & no                   & single               & yes    & no      & no  \\
MotionBase           & 1M    & -     & -                & yes                  & multi                & yes    & yes     & yes \\ \hline
MotionMillion (ours) & 2M    & \textgreater{}2000     & \textgreater{}20 & no                   & multi                & yes    & yes     & yes \\ \hline
\end{tabular}%
}
\caption{Statistics comparison between our MotionMillion and other motion datasets.}
\label{tab:comp_curr_dataset}
\end{table}

\begin{figure}[t]
    \centering
    \includegraphics[width=\linewidth]{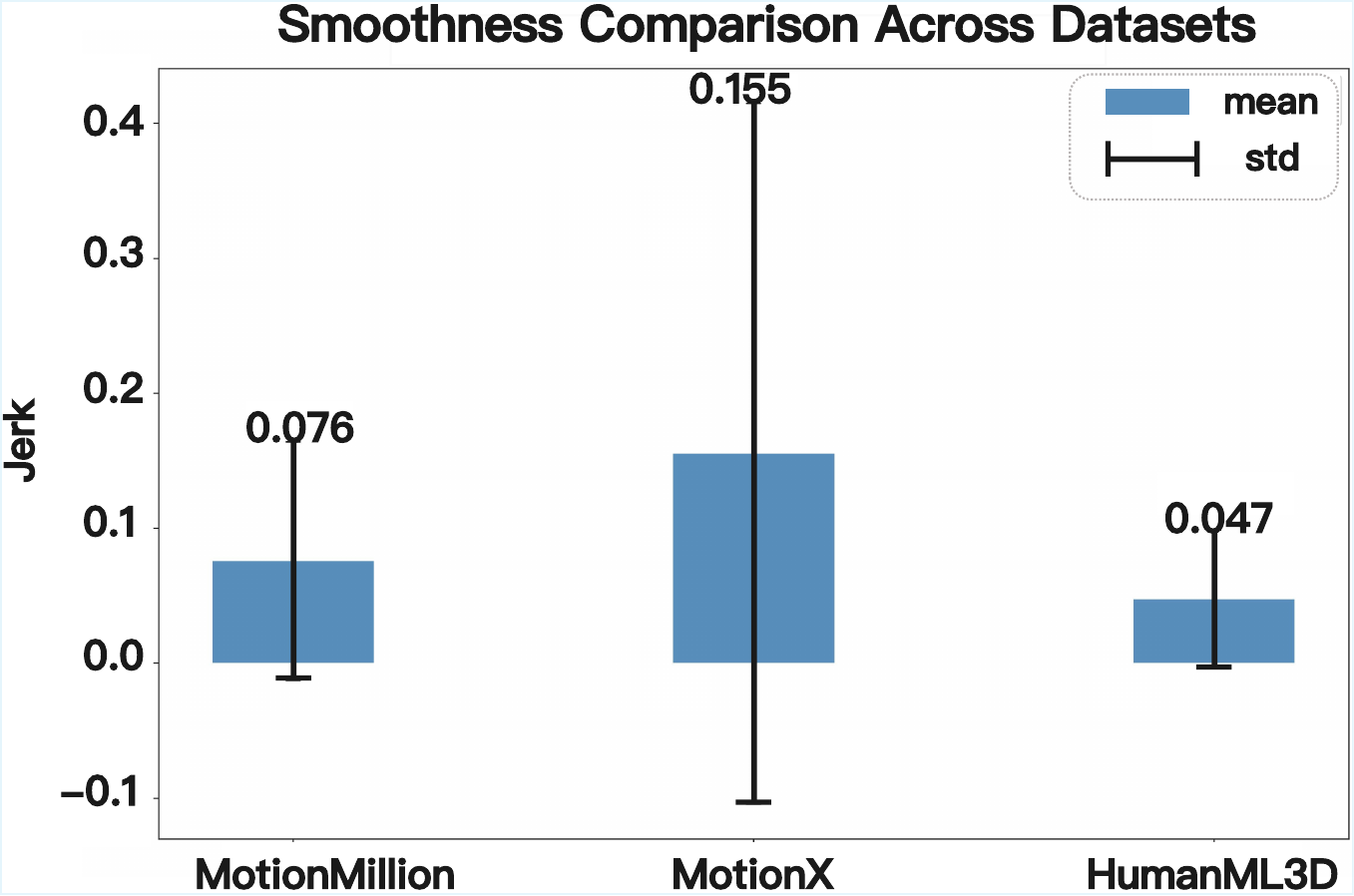}
    \caption{Jerk comparison across MotionMillion, MotionX, and HumanML3D. Our MotionMillion exhibits the lowest jerk values, indicating that it produces smoother motion.}
    \label{fig:data_analysis_2}
\end{figure}

As presented in Table~\ref{tab:comp_curr_dataset}, our collected MotionMillion dataset comprises over 2,000 hours of human motion clips, encompassing more than 2 million motion sequences. Each sequence is ensured to exceed one second in duration, recorded at a frame rate of 30 frames per second.
To enhance both the appearance and motion diversity, we curate a large collection of monocular videos from online sources, capturing a wide range of real-world scenarios. As depicted in Figure~\ref{fig:data_analysis}, our dataset encompasses various real-life settings, including indoor activities (e.g., performances, boxing) and outdoor movements (e.g., exercise, martial arts).

We further assess data quality from multiple perspectives, including pose diversity, semantic diversity (illustrated at the top of Figure~\ref{fig:data_analysis}), and motion smoothness (depicted in Figure~\ref{fig:data_analysis_2}). To evaluate motion smoothness, we compute the average jerk of the dataset, as formulated in Equation~\ref{eq:jerk}. The results indicate that the motion smoothness achieved through our data construction pipeline is significantly enhanced compared to MotionX and near to the overall smoothness of HumanML3D, thereby ensuring the high quality of our dataset.

Additionally, we visualize the top-2 principal components of body part poses and text features using t-SNE dimensionality reduction. The comparative analysis of pose distributions demonstrates that the diversity of poses within MotionMillion is on par with that of other datasets. However, in terms of semantic diversity, our dataset exhibits a significantly richer distribution compared to existing benchmarks. This is an expected outcome, as human motion often involves recurring pose patterns, yet distinct pose combinations yield semantically varied movements.
Consequently, the increased semantic richness in our dataset presents challenges for smaller models to fully capture the underlying distribution, thereby stimulating further research in text-to-motion generation. In particular, our dataset underscores the necessity of scaling up model sizes and advancing the field towards zero-shot human motion generation.

\begin{figure}
    \centering
    \includegraphics[width=0.7\linewidth]{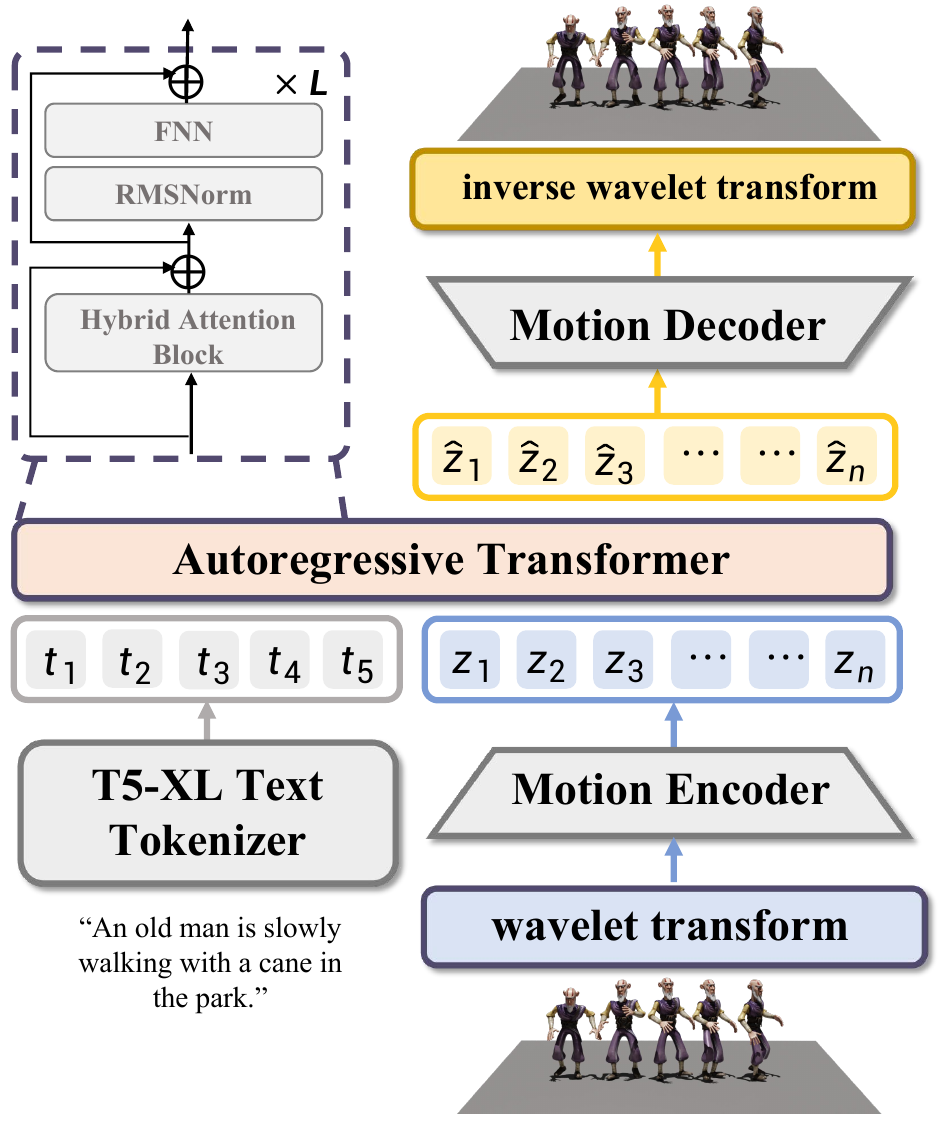}
    \caption{Overview of our scalable model architecture, which utilize FSQ as a motion tokenizer and an autoregressive transformer to generate the motion from the given text.}
    \label{fig:framework}
\end{figure}

\section{Architecture}
\label{sec:architecture}
Building upon the constructed large-scale annotated motion dataset, we aim to train a foundational motion generation model capable of zero-shot motion generalization, especially the ability to generate complex compositional motions. Inspired by the successful scaling strategies observed in natural language and computer vision\citep{rombach2022high,gpt3,gpt4,sora}, we adopt a discrete autoregressive architecture, as illustrated in Fig.~\ref{fig:framework}.
The proposed model consists of two key stages: Efficient Motion Tokenization and Scalable Motion Generation.
1)	Efficient Motion Tokenization: This stage employs a finite scalar quantizer (FSQ) to learn discrete representations of human motion sequences, enabling efficient encoding of continuous motion data into a compact and structured format.
2) Scalable Motion Generation: Leveraging the LLAMA architecture, the model takes text inputs as prompts and scales from 1B to 7B parameters, facilitating high-capacity motion generation with strong generalization capabilities.
This design enables the model to generate realistic and contextually coherent human motion sequences while maintaining scalability.

\subsection{Motion Representations}
We begin by introducing our motion representation. To mitigate errors introduced by the inverse kinematics process in the HumanML3D format while preserving redundant information (e.g., velocity), we reformulate and refine the motion representation \(x^i\) in a manner consistent with previous work on character control~\cite{starke2019neural, starke2022deepphase, shi2024interactive}. Specifically, the \(i\)-th pose \(x^i\) is defined as a tuple comprising: root linear velocities \((\dot{r}^x, \dot{r}^z \in \mathbb{R})\) on the XZ-plane, root angular velocity \(\dot{r}^a \in \mathbb{R}^6\) represented in 6D rotations, local joint positions \(p^i \in \mathbb{R}^{3N}\), local velocities \(v^i \in \mathbb{R}^{3N}\), and local rotations \(r^i \in \mathbb{R}^{6N}\) relative to the root space, where \(N\) denotes the number of joints. Formally, this is expressed as:
\[
x^i = \{\dot{r}^x,\dot{r}^z,\dot{r}^a,\,p^i,\,v^i,\,r^i\}.
\]

A significant advantage of our representation is that it eliminates the need for an inverse kinematics process to obtain SMPL or BVH representations, as required in previous approaches. Moreover, our representation can be losslessly converted to relative rotations akin to those in SMPL. Additionally, because both the rotation and position components are derived from the same skeletal structure, they provide mutual regularization. Notably, the rotation component in the HumanML3D format is erroneous~\cite{humanmlissue}, which heavily hinders the applications of downstream tasks. Rather than discarding this flawed rotation component as done in~\cite{meng2024rethinking}, we undertake engineering corrections to rectify it. We hope our representation could correct previous mistakes and guide future development.

\subsection{Efficient Motion Tokenization}
We leverage Finite Scalar Quantization (FSQ), a codebook-free quantization mechanism that replaces distance-based codebook matching with deterministic discretization, enabling scalable and stable representation learning~\citep{lu2024scamo}.  

The vanilla FSQ architecture operates in three key steps. First, the latent vector z, produced by the motion encoder, is normalized via a sigmoid function to constrain its values within a bounded range \([0, 1]\). Next, each dimension of the normalized latent is discretized into \(L\) uniformly spaced integers using a rounding operation:  
\vspace{-0.5em}
\begin{equation}
\hat{\mathbf{z}} = \mathcal{Q}(\mathbf{z}) = \mathtt{round}\left(f(\mathbf{z}) \cdot (L-1)\right),
\end{equation},
where \(L\) defines the number of quantization levels per dimension. This produces a discrete code \(\hat{\mathbf{z}} \in \{0, 1, \dots, L-1\}^d\), with \(d\) denoting the latent dimension. Unlike VQ-VAE’s explicit codebook, FSQ implicitly defines a structured grid of \(L^d\) unique codes, eliminating the need to store physical embeddings while ensuring full code utilization.  

The model is optimized solely via reconstruction loss:  
\begin{equation}
\mathcal{L} = \|\mathbf{m} - \mathtt{Dec}(\mathbf{z}_q)\|_2^2,
\end{equation}
removing auxiliary losses (e.g., codebook commitment terms) required in VQ-VAE.  
This structured quantization paradigm provides a robust alternative to traditional VQ for representing complex motion data at scale.

Unlike the vanilla FSQ method, as shown in Fig.~\ref{fig:framework}, we first use a wavelet transform before inputting the motion into the motion encoder, and after the motion decoder, we use the inverse wavelet transform to restore the motion. This can largely suppress the jitter problem caused by the information loss of discrete encoding.

\subsection{Scalable Motion Generation}
To effectively scale our model, we adopt a transformer-decoder-based architecture. The framework first compresses and encodes motion using a finite scalar quantizer (FSQ), while textual information is processed using the T5-XL large language model, which performs word-level encoding. The encoded motions and texts are denoted as $\{m_i\}_{i=1}^{n}$ and $\{T_i\}_{i=1}^{w}$, where $n$ and $w$ represent the number of encoded motion tokens and the word tokens.

Unlike standard causal attention mechanisms used in transformers, our approach employs a mixed attention strategy, where attention among words is bidirectional, whereas attention between motion sequences remains causal. As illustrated in Fig.~\ref{fig:framework}, the encoded text and motion representations are fed into a series of stacked Hybrid Attention Blocks (HABs), with the final output directed to a classification head. Each HAB consists of:
1) Two RMS-Norm layers, which significantly mitigates the training instability. 2) One mixed attention module, which capture intricate relationships between text and motion features, and 3) one feedforward network(FFN) module, which could fuse the feature from a channel aspect. To optimize the model, we apply cross-entropy loss to the logits produced by the final classification head, formulated as:
\begin{equation}
    \mathcal{L} = - \sum_{i=1}^{n} \log p(\hat{m}_i | m_{<i}, T_1,...,T_w)
\end{equation}
This approach ensures robust motion-text alignment while maintaining scalability and training stability.

\begin{table}[t]
\resizebox{\columnwidth}{!}{%
\begin{tabular}{cccc}
\hline
\multirow{2}{*}{Method} & \multicolumn{3}{c}{Dataset}                                                  \\ \cline{2-4} 
                        & \multicolumn{1}{l}{HumanML3D} & \multicolumn{1}{l}{MotionX} & MotionMillion \\ \hline
ScaMo~\citep{lu2024scamo}             & 63.3                             & 84.1                           & 88.9             \\
Ours                    & \textbf{41.9}                             & \textbf{57.4}                       & \textbf{45.5}             \\ \hline
\end{tabular}%
}
\vspace{-1em}
\caption{MPJPE of reconstruction comparison across different datasets, where ScaMo's FSQ model and ours are trained on MotionUnion and MotionMillion, respectively.}
\label{tab:recons_comp}
\end{table}
\begin{table}[t]
\centering
\resizebox{0.85\columnwidth}{!}{%
\begin{tabular}{cccc}
\hline
\textbf{}    & MPJPE$\downarrow$ & Mean Acc$\downarrow$   & Max Acc$\downarrow$     \\ \hline
GT           & -     & 2.0          & 9.0           \\
w/o wavelet & 46.8    & 6.0          & 15.0          \\
w/ wavalet   & \textbf{45.5}    & \textbf{4.0} & \textbf{12.0} \\ \hline
\end{tabular}%
}
\caption{Ablation on whether to use wavelet transformation during training the FSQ model, where Acc represents the acceleration.}
\label{ablation:wavelet}
\end{table}
\begin{table}[t]
\centering
\resizebox{0.7\columnwidth}{!}{%
\begin{tabular}{lcccc}
\hline
Method & FID$\downarrow$ & R@1$\uparrow$ & R@2$\uparrow$ & R@3$\uparrow$ \\ \hline
ScaMo     & 89.0 & 0.67 & 0.81 & 0.87 \\ \hline
Ours-1B   & 31.3 & 0.74 & 0.87 & 0.92 \\ 
Ours-3B   & 10.8 & 0.79 & \textbf{0.91} & 0.94 \\
Ours-7B   & \textbf{10.3} & 0.79 & 0.90 & 0.94 \\ \hline
\end{tabular}%
}
\caption{Quantitative comparison of ScaMo and our models of different sizes on MotionMillion.}
\label{tab:comp_on_motionmillion}
\end{table}

\section{Experiments}
\label{sec:experiments}

\begin{figure*}[!t]
    \centering
    \includegraphics[width=1\textwidth]{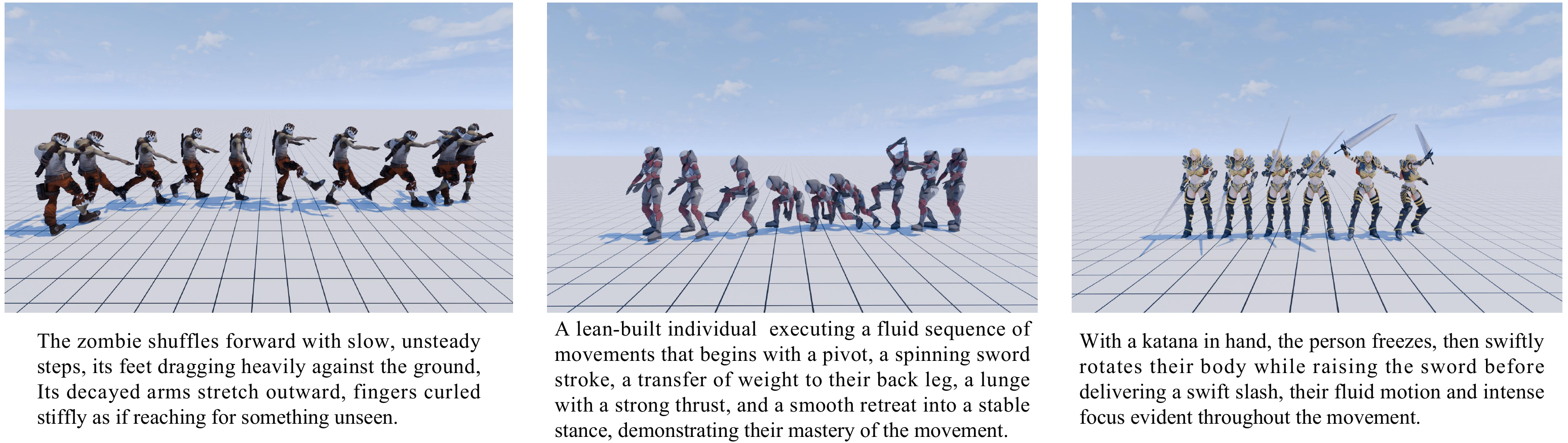}
    \vspace{-2em}
    \caption{Our model demonstrates robust performance in generating coherent motions from complex compositional textual descriptions.}
    \vspace{-1em}
    \label{fig:generation_result}
\end{figure*}

\subsection{FSQ Reconstruction Comparison}

\fk{\textbf{Reconstruction Comparison with Different Datasets.}}
We first compare the reconstruction performance of different datasets after conducting FSQ training with the same number of parameters on both our dataset and the MotionUnion dataset. As shown in Tab.~\ref{tab:recons_comp}, our FSQ model is trained on the training set of MotionMillion, while the FSQ model from Scamo is trained on the MotionUnion dataset. We observe that the FSQ model trained on the MotionMillion training set, using the same number of parameters, achieves the best performance across HumanML3D, MotionX, and MotionMillion.

\fk{\noindent \textbf{Ablation Study Wavelet Transformation.}}
\fk{As shown in the Tab.~\ref{ablation:wavelet}, we exhibit the reconstruction results via MPJPE metrics and jitter by calculating the acceleration of the reconstructed motion. 

We find that training vanilla FSQ (a.k.a. w/o wavelet) leads to a large deviation between the acceleration of the motion reconstructed by the model and that of the ground truth (GT), which means there is a significant data jitter. We analyze that this may be because the discretized compression inevitably causes information loss, resulting in a deviation in the modeling of high-frequency information, thus leading to severe jitter. Therefore, we use the wavelet transform to transform the input motion, and after using the wavelet transform, the deviation between the acceleration of the motion reconstructed by the model and that of the GT is significantly reduced. At the same time, the MPJPE of the model also gains a little better improvement.}

\subsection{Quantitative Comparison on Different Models}

\fk{We followed the previous method~\cite{humantomato}, and trained an evaluator on our dataset to assess the impact of different model scales on the generation ability.}
\fk{As shown in Tab.~\ref{tab:comp_on_motionmillion}, we can find that our model significantly surpasses the ScaMo performance. Besides, with the increase of the model size, FID and R-precision are gradually improved. However, we found that when the model size improved from 3B to 7B, it was difficult to effectively reflect the difference between the two models from the metrics. This is also in line with the current dilemma of generative models, that is, metrics often can not fully reflect the capabilities of the model. Therefore, in the following sections, we further propose the MotionMillion-Eval benchmark to evaluate zero-shot generation capabilities of different models.}

\subsection{Zero-shot Potential Verification}
\textbf{Benchmark and Metric.}
We introduce MotionMillion-Eval, a novel benchmark designed to assess the quality of text-to-motion generation models through human verification. This benchmark comprises 126 prompts derived from real-world industrial standards, covering various motion categories, including daily life, work, arts, communication, combat, sports, dance, and nonhuman behavior.
We evaluate different text-to-motion models with three human evaluation dimensions on MotionMillion-Eval:
1) Text Alignment, 2) Motion Smoothness, and 3) Physical Plausibility. The scoring details for the three dimensions are provided in the supplementary materials.

\begin{table}[t]
\resizebox{\columnwidth}{!}{%
\begin{tabular}{lccc}
\hline
Model & \multicolumn{1}{l}{Text Alignment} & \multicolumn{1}{l}{Physical Plausibility} & \multicolumn{1}{l}{Motion Smoothness} \\ \hline
MDM~\citep{mdm}      & 195.5        & 478.5        & 416.5        \\
MotionGPT~\citep{motiongpt}     & 170          & 497          & 501.5        \\
T2M-GPT~\citep{t2mgpt} & 207          & 495.5        & 500          \\
ScaMo-3B~\citep{lu2024scamo}    & 226.6        & 477.5        & 494          \\ \hline
Ours-1B       & 170.3        & \textbf{497} & \textbf{501} \\
Ours-3B       & 238.6        & 496          & 499.5        \\
Ours-7B       & \textbf{261} & 495.5        & \textbf{501} \\ \hline
\end{tabular}%
}
\caption{Average human evaluation results under the aspect of Text Alignment, Physical Plausibility, and Motion Smoothness on MotionMillion-Eval between Different Models.}
\label{tab:comp_metrics}
\end{table}
\begin{table}[t]
\resizebox{\columnwidth}{!}{%
\begin{tabular}{lccc}
\hline
\textbf{Ours-7B vs. ScaMo-3B (win/tie/lose)} & Annotator 1 & Annotator 2 & Annotator 3 \\ \hline
Overall (126)                           & \cellcolor{green!20}45/49/32    & \cellcolor{green!20}32/76/18    & \cellcolor{green!20}47/53/26    \\
\hline
Art/Dance (2)                      & \cellcolor{yellow!20}1/0/1       & \cellcolor{yellow!20}0/2/0       & \cellcolor{green!20}1/0/1       \\
Combat (6)                         & \cellcolor{green!20}3/2/1       & \cellcolor{yellow!20}2/2/2       & \cellcolor{green!20}3/2/1       \\
Communication (20)                 & \cellcolor{green!20}5/13/2      & \cellcolor{green!20}3/16/1      & \cellcolor{green!20}8/11/1      \\
Daily life (74)                    & \cellcolor{green!20}24/27/23    & \cellcolor{green!20}21/40/13    & \cellcolor{green!20}25/31/18    \\
Non-human behavior (2)             & \cellcolor{green!20}2/0/0       & \cellcolor{yellow!20}0/2/0       & 0/1/1       \\
Sports (16)                        & \cellcolor{green!20}6/6/4       & \cellcolor{green!20}5/9/2       & \cellcolor{green!20}7/6/3       \\
Work (6)                           & \cellcolor{green!20}4/1/1       & \cellcolor{green!20}1/5/0       & \cellcolor{green!20}3/2/1       \\ \hline
\end{tabular}%
}
\caption{Detailed comparison between ours-7B and ScaMo-3B. The green cells indicate that our model outperforms competing approaches, the yellow cells represent a tie, and the white cells denote that our model underperforms.}
\label{tab:comp_detail_bench}
\end{table}

\noindent\textbf{Comparisons Between Different Models.} As shown in Tab.~\ref{tab:comp_metrics},
We evaluate multiple models on MotionMillion-Eval. Compared with the ScaMo-3B model, our 3B model significantly outperforms ScaMo in terms of Text Alignment, Physical Plausibility, and Motion Smoothness. This indicates that compared to the MotionUnion dataset used for training ScaMo, the model trained with our dataset has stronger generalization performance. In addition, as the model scale increases (ranging from 1B to 7B), the effect in the dimension of Text Alignment is significantly improved. This shows that the scale and diversity of our data can effectively support the expansion of the model scale, thus better paving the way for realizing zero-shot applications. 

Besides, we find that for the two metrics of physical feasibility and motion smoothness, expanding the model size does not bring a significant increase. We analyze that the reason for this is that these two metrics do not need to measure the alignment degree between motion and text, but only need to focus on the quality of motion itself. Therefore, we believe that these two metrics are highly correlated with the motion quality of the dataset itself. Further, when we compare the MDM, T2M-GPT, and ScaMo-3B models, we find that our model can be comparable to the former two in terms of Physical Plausibility and Motion Smoothness, and is significantly better than the ScaMo-3B model. Therefore, this further shows that the quality of MotionMillion we constructed can significantly reach almost the same level as HumanML3D and is significantly better than MotionX.

\noindent\textbf{Detailed Comparisons Between Ours-7B and ScaMo-3B Models.}
We compared the generative capabilities of both models on diverse instruction categories, with three professional annotators voting on the better outputs for identical text prompts. As shown in Table~\ref{tab:comp_detail_bench}, our model matches ScaMo-3B in the Art/Dance and Non-human behavior subsets but surpasses it in all other areas, demonstrating the enhanced diversity of motions achievable by our dataset.

\subsection{Qualitive Results}
We further present visualization results, as shown in Fig.~\ref{fig:generation_result}. Our model demonstrates: 1) the ability to comprehend abstract concepts, effectively recognizing and generating the characteristic walking posture of a zombie; and 2) strong instruction-following capabilities, accurately interpreting long text descriptions and producing corresponding combined motions. These results indicate that MotionMillion, the largest-scale dataset we propose, has the potential to advance text-to-motion generation into the zero-shot era.
\section{Conclusion}
In this paper, we take the first step toward advancing human motion generation into the zero-shot era. We begin by analyzing the limited generalization ability of existing methods, attributing it to the constrained size of current datasets. To address this, we propose an efficient data annotation mechanism, establishing the largest annotated human motion dataset. Furthermore, to assess the zero-shot capabilities, we introduce MotionMillion-Eval, a dedicated evaluation benchmark. Building on this foundation, we successfully scale our model to 7B parameters using a scalable architecture, achieving state-of-the-art performance on the benchmark and demonstrating strong zero-shot capabilities. We believe this work lays a crucial foundation for advancing zero-shot applications in motion generation.

{
    \small
    \bibliographystyle{ieeenat_fullname}
    \bibliography{main}
}

\clearpage
\setcounter{page}{1}
\maketitlesupplementary

\section{Overview}
In the following, we provide additional implementation details, including the data distribution and prompts in Sec.~\ref{supp:ddp}, finegrained scoring criteria in MotionMillion-Eval in Sec.~\ref{supp:eval_detail}, and 126 prompts used by MotionMillion-Eval in Sec.~\ref{supp:prompts}.
Various generations of out-domain and complex compositional long motions are shown in our demo video.

\section{Data Distribution and Prompts}
\label{supp:ddp}
\begin{figure}[t]
    \centering
    \includegraphics[width=\linewidth]{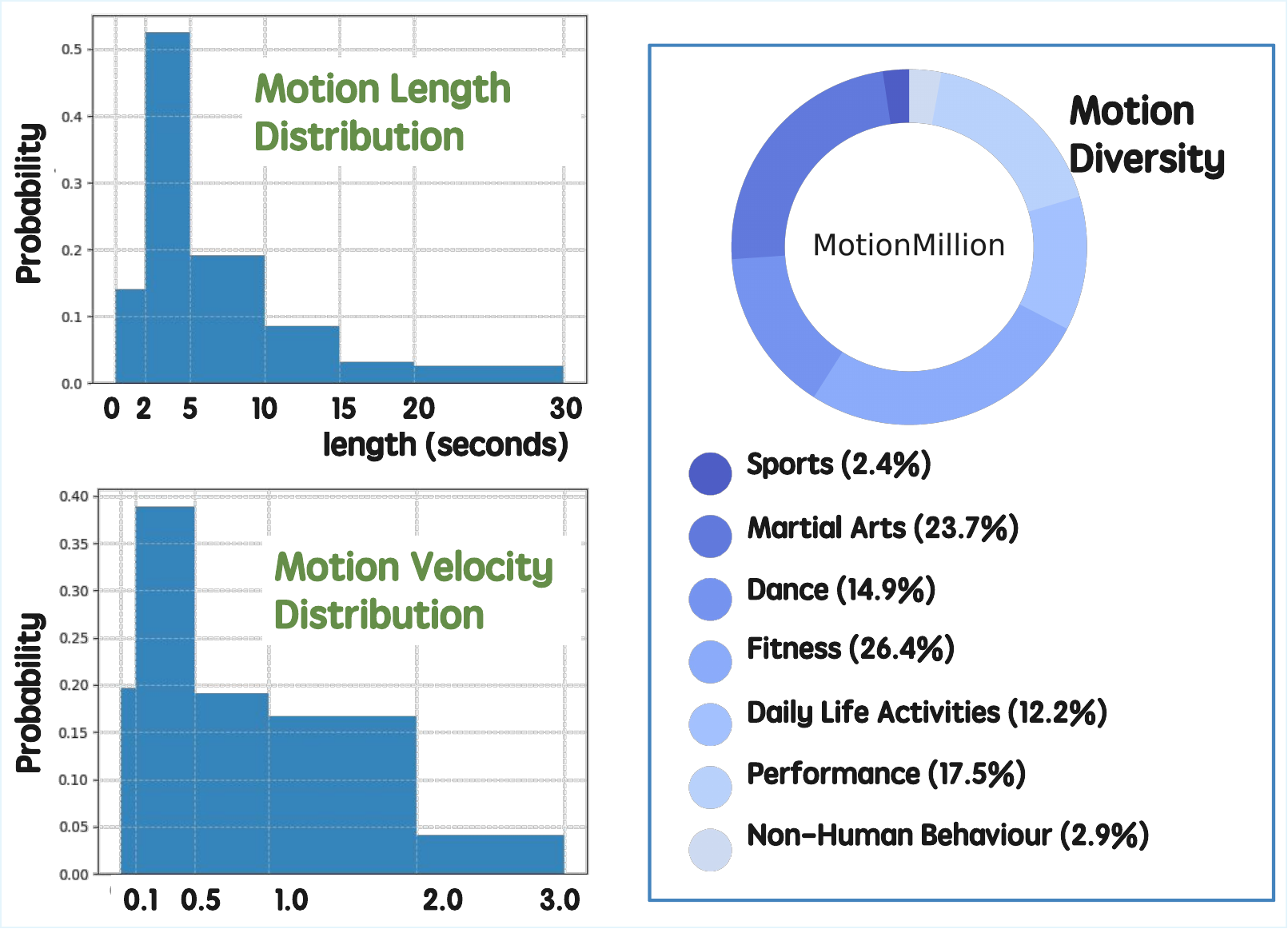}
    \caption{Data Distributions of MotionMillion.}
    \label{fig:supp_datadis}
\end{figure}

\begin{figure}[t]
    \centering
    \includegraphics[width=\linewidth]{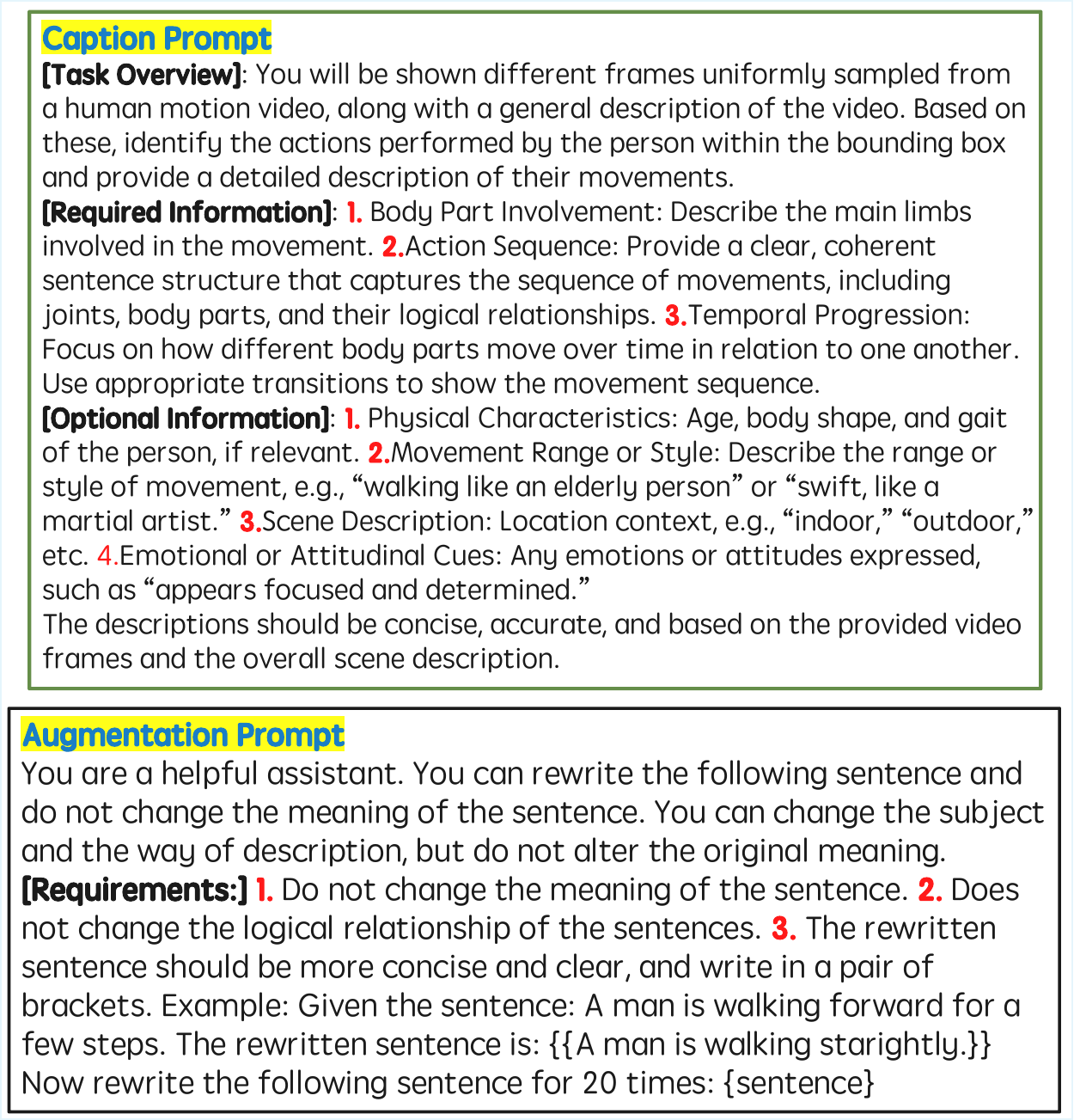}
    \caption{Prompt used during captioning the motions in the web-scale human videos and text rewrite in inference stage.}
    \vspace{-2em}
    \label{fig:supp_prompt}
\end{figure}

We further demonstrate the data distribution of motion length, motion velocity, and motion diversity in Fig.~\ref{fig:supp_datadis}. We also provide the prompt used during captioning the motions in the web-scale human videos and text rewrite in the inference stage in Fig.~\ref{fig:supp_prompt}.

\section{Scoring Criteria Details of MotionMillion-Eval}
\label{supp:eval_detail}
The scoring criteria details for each dimension (Text Alignment, Motion Smoothness, and  Physical Plausibility) are defined as follows:

\textbf{Text Alignment (TA).}Score = 4: The generated motion is fully aligned with the textual prompt, accurately depicting all specified elements and details.
Score = 3: The motion generally corresponds to the prompt, though minor discrepancies may be present in certain details.
Score = 2: The motion exhibits clear misalignment with the prompt, with significant omissions or deviations from the described content.
Score = 1: The generated motion is entirely inconsistent with the prompt, displaying substantial inaccuracies in key scenes or actions.

\textbf{Motion Smoothness (MS).} Score = 4: The motion is highly fluid and natural, with smooth and seamless transitions between movements. Score = 3: The motion is generally smooth, though minor unnatural artifacts may occasionally appear in specific segments. Score = 2: The motion lacks fluidity, exhibiting noticeable discontinuities or stuttering. Score = 1: The motion appears highly unnatural, with frequent stuttering and abrupt transitions that disrupt coherence and comprehensibility.

\textbf{Physical Plausibility (PP).} Score = 4: The generated motion adheres to real-world physical laws, accurately simulating object interactions, lighting, shadows, and collision effects.
Score = 3: Multiple instances of physically implausible motion, lighting inconsistencies, or unrealistic interactions are observed, though the primary actions maintain a degree of coherence.
Score = 2: The generated motion exhibits substantial violations of physical laws, with unrealistic object interactions or lighting effects that diminish realism.
Score = 1: The motion is entirely implausible, featuring severe distortions in object dynamics, lighting, or interactions, making the scene difficult to interpret.

\begin{figure*}
    \centering
    \includegraphics[width=0.8\linewidth]{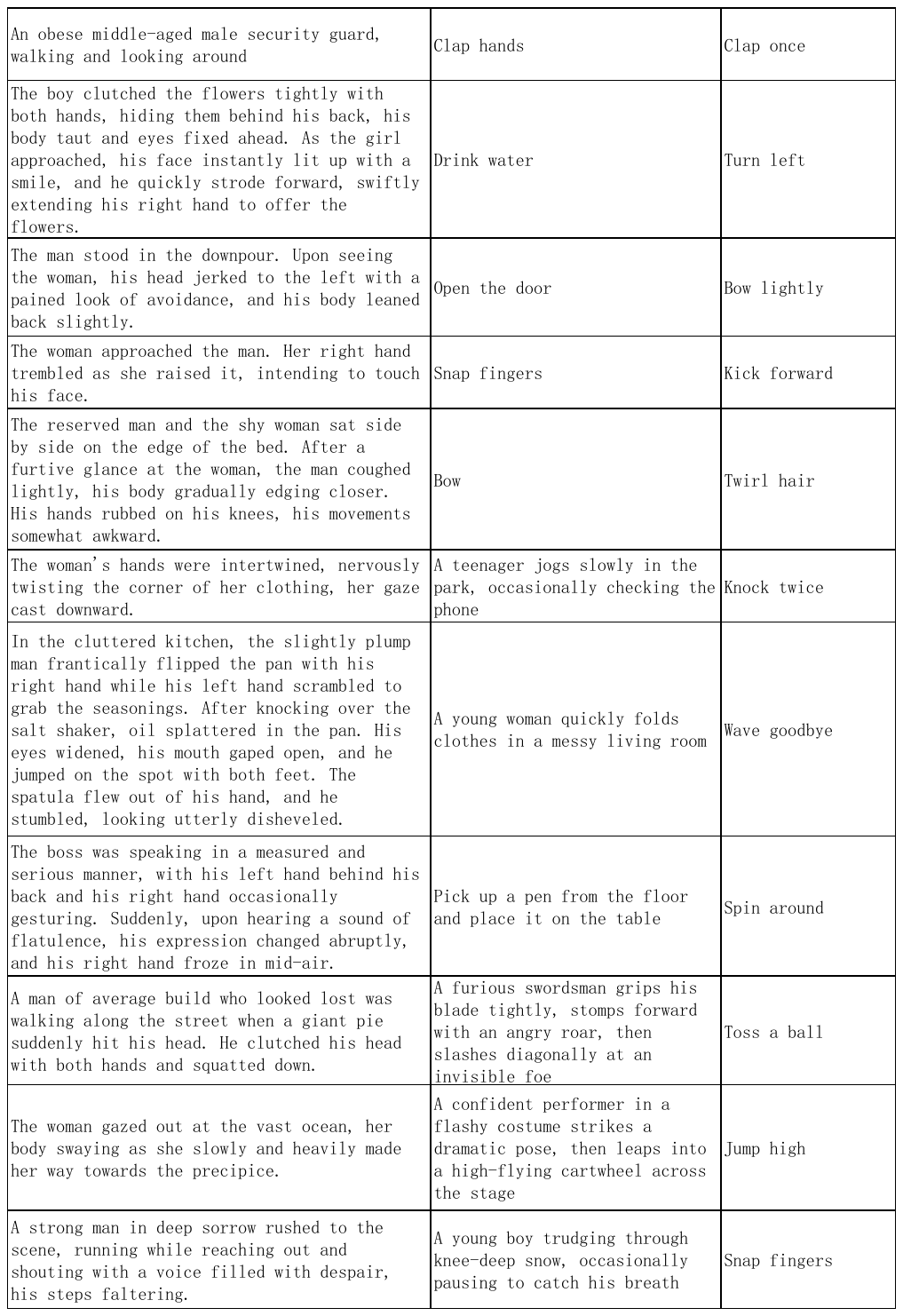}
\end{figure*}

\begin{figure*}
    \centering
    \includegraphics[width=0.8\linewidth]{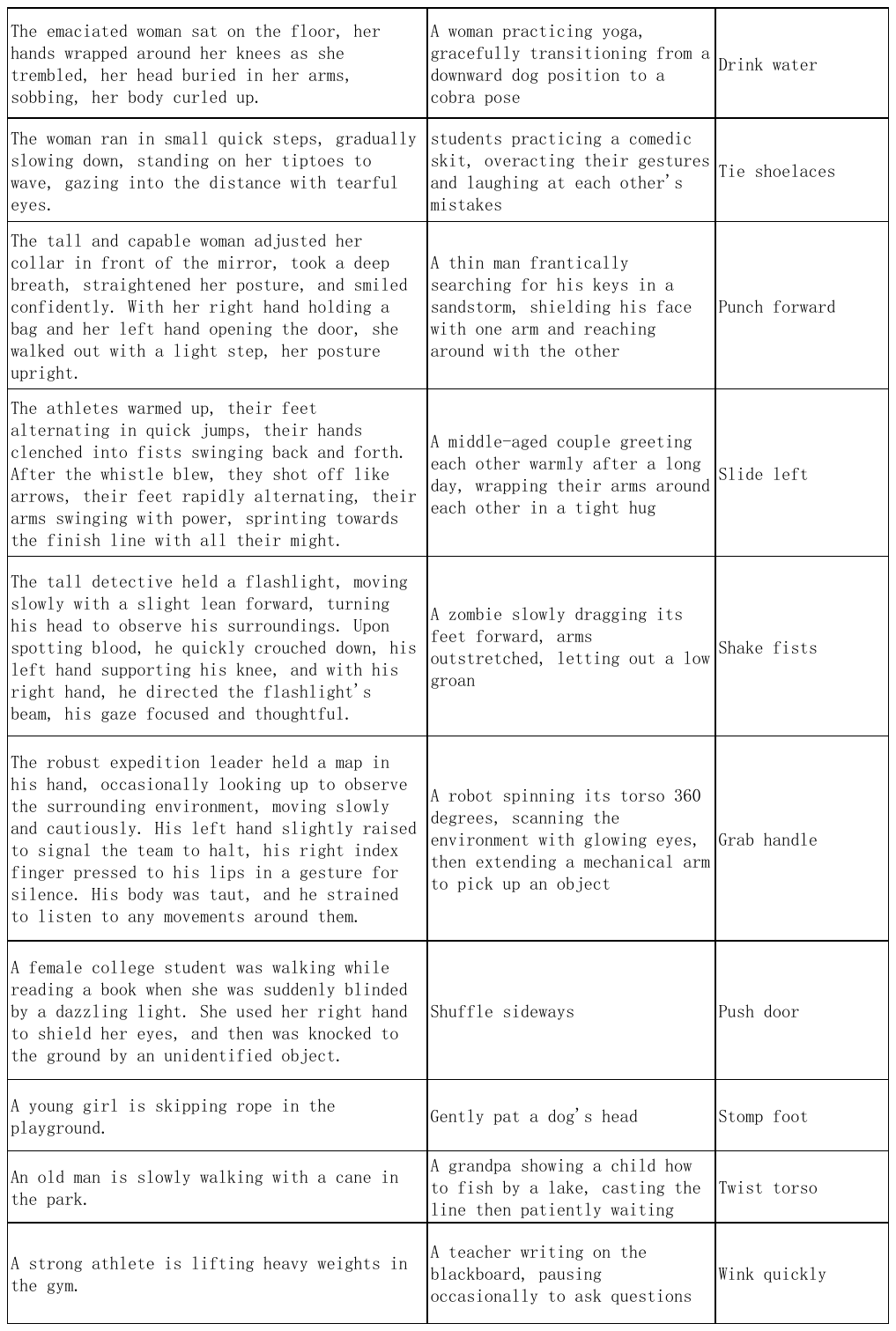}
\end{figure*}

\begin{figure*}
    \centering
    \includegraphics[width=0.8\linewidth]{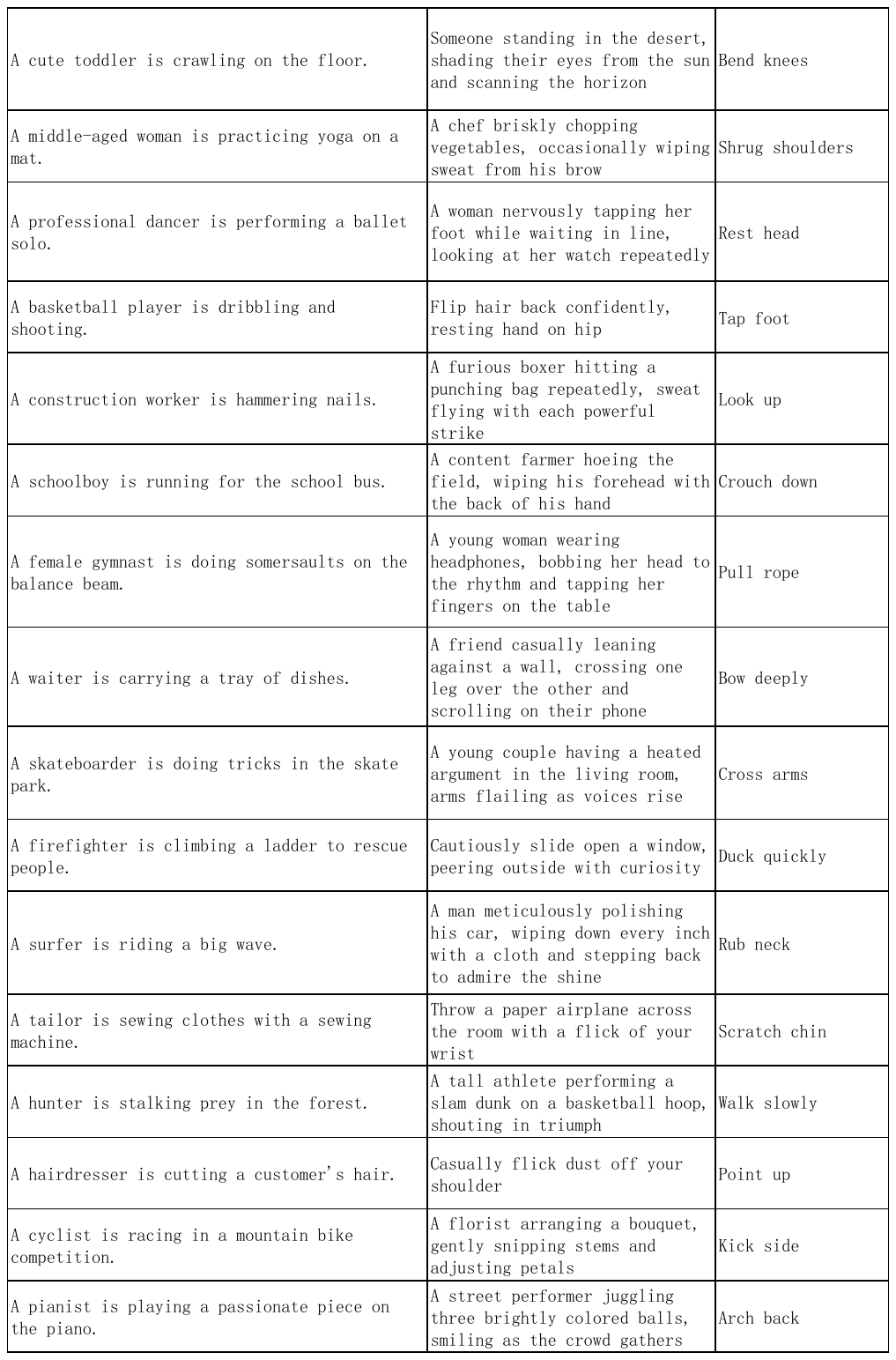}
\end{figure*}

\begin{figure*}
    \centering
    \includegraphics[width=0.8\linewidth]{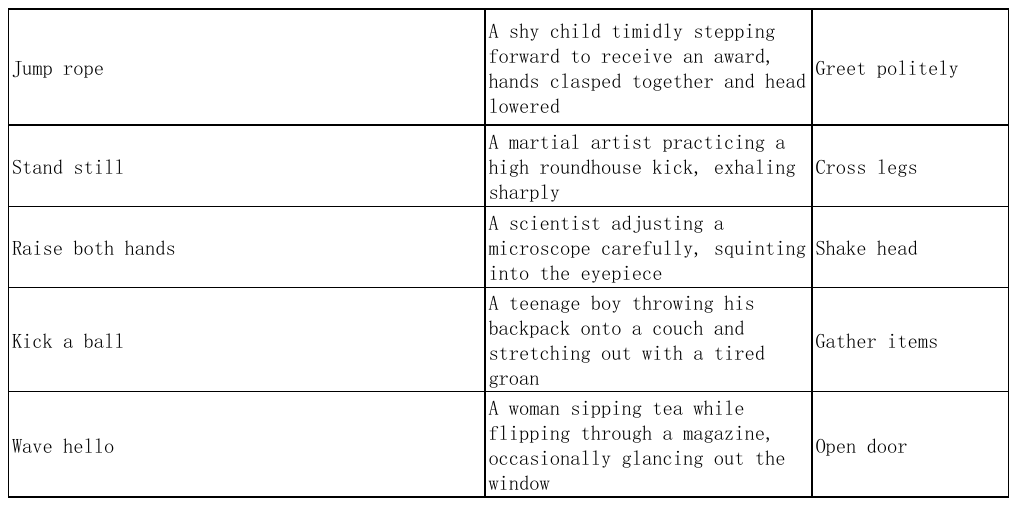}
\end{figure*}
\section{Prompts in MotionMillion-Eval}.
\label{supp:prompts}
We show all the prompts in the figures below.

\end{document}